\documentclass{article}
\usepackage{ijcai26}

\usepackage{times}
\usepackage{soul}
\usepackage{url}
\usepackage[hidelinks]{hyperref}
\usepackage[utf8]{inputenc}
\usepackage[small]{caption}
\usepackage{graphicx}
\usepackage{amsmath}
\usepackage{amsthm}
\usepackage{booktabs}
\usepackage{algorithm}
\usepackage{algorithmic} 
\usepackage[switch]{lineno}
\usepackage{tabularx}
\usepackage{makecell}
\usepackage{array}
\usepackage{caption}
\usepackage{colortbl}
\usepackage{tabularx}
\usepackage{multirow}
\usepackage{multicol}
\usepackage{graphicx}
\usepackage{blindtext}
\usepackage{hyperref}
\usepackage{xcolor, colortbl}
\usepackage{relsize}
\usepackage{CJKutf8}
\usepackage{pdfpages}
\usepackage{subcaption}
\usepackage{adjustbox}
\usepackage{amssymb}
\usepackage{listings}
\usepackage{tcolorbox}
\usepackage{cleveref}
\usepackage{fancyhdr}
\usepackage[table,xcdraw,dvipsnames]{xcolor}
\usepackage{caption}
\usepackage{multirow}
\usepackage{graphicx}
\usepackage[table]{xcolor}
\usepackage{adjustbox}
\usepackage{caption}
\usepackage{multirow}   
\usepackage{graphicx}   
\usepackage{adjustbox}  
\usepackage{pifont}
\usepackage{graphicx}
\usepackage{float}
\usepackage{amsthm}
\usepackage{booktabs}

\usepackage[T1]{fontenc}
\usepackage[utf8]{inputenc}
\newcounter{RZNumberOfComments}
\stepcounter{RZNumberOfComments}

\newcommand{\lcite}[1]{\textcolor{blue}{cite here}}

\newcommand{\cmark}{\ding{51}}
\newcommand{\xmark}{\ding{55}}

\usepackage{enumitem}

\title{Med-StepBench: A Hierarchical Reasoning Framework for Evaluating Hallucinations in Medical Vision-Language Models}

\author{
Minh Khoi Nguyen$^1$\thanks{Equal contribution.}
\and
Dai Lam Le$^1$\footnotemark[1]
\and
Amir Reza Jafari$^2$
\and
Tuan Dung Nguyen$^1$ 
\and
\\Mai Hong Son$^3$
\and
Mai Huy Thong$^3$
\and
Quang Huy Nguyen$^4$
\and
Thanh Trung Nguyen$^3$
\and
\\Reza Farahbakhsh$^2$
\and
Noel Crespi$^2$\thanks{Corresponding authors.}
\and
Phi Le Nguyen$^1$\footnotemark[2]\\[4pt]
\affiliations
$^1$AI4LIFE, Hanoi University of Science and Technology, Vietnam\\
$^2$SAMOVAR, T\'el\'ecom SudParis, Institut Polytechnique de Paris, France\\
$^3$108 Military Central Hospital, Vietnam 
$^4$Hanoi Medical University\\
}

\begin{document}
\maketitle
\begin{abstract}
Large vision-language models (VLMs) demonstrate strong performance in medical image understanding, but frequently generate clinically plausible yet incorrect statements, raising significant safety concerns. Existing medical hallucination benchmarks primarily focus on 2D imaging with one-shot diagnostic questions, offering limited insight into whether predictions are grounded in correct localization and abnormality identification, allowing critical reasoning errors to remain hidden behind seemingly correct diagnoses. We introduce \textit{Med-StepBench}, the first large-scale benchmark for step-wise hallucination detection in 3D oncological PET/CT, comprising over 12,000 images and more than 1,000,000 image–statement pairs across volumetric and multi-view 2D data, which decomposes clinical reasoning into four expert-designed diagnostic stages. Using clinician-verified annotations, we perform the first step-level evaluation of general-purpose and medical VLMs, revealing systematic failure modes obscured by aggregate accuracy metrics. Furthermore, we show that current VLMs are highly susceptible to adversarial yet clinically plausible intermediate explanations, which significantly amplify hallucinations despite contradictory visual evidence. Together, our findings highlight fundamental limitations in grounding multi-step clinical reasoning and establish \textit{Med-StepBench }as a rigorous benchmark for developing safer and more reliable medical VLMs.
\end{abstract}

\section{Introduction}

\begin{figure}[!t]
    \centering
    \includegraphics[width=0.92\linewidth]{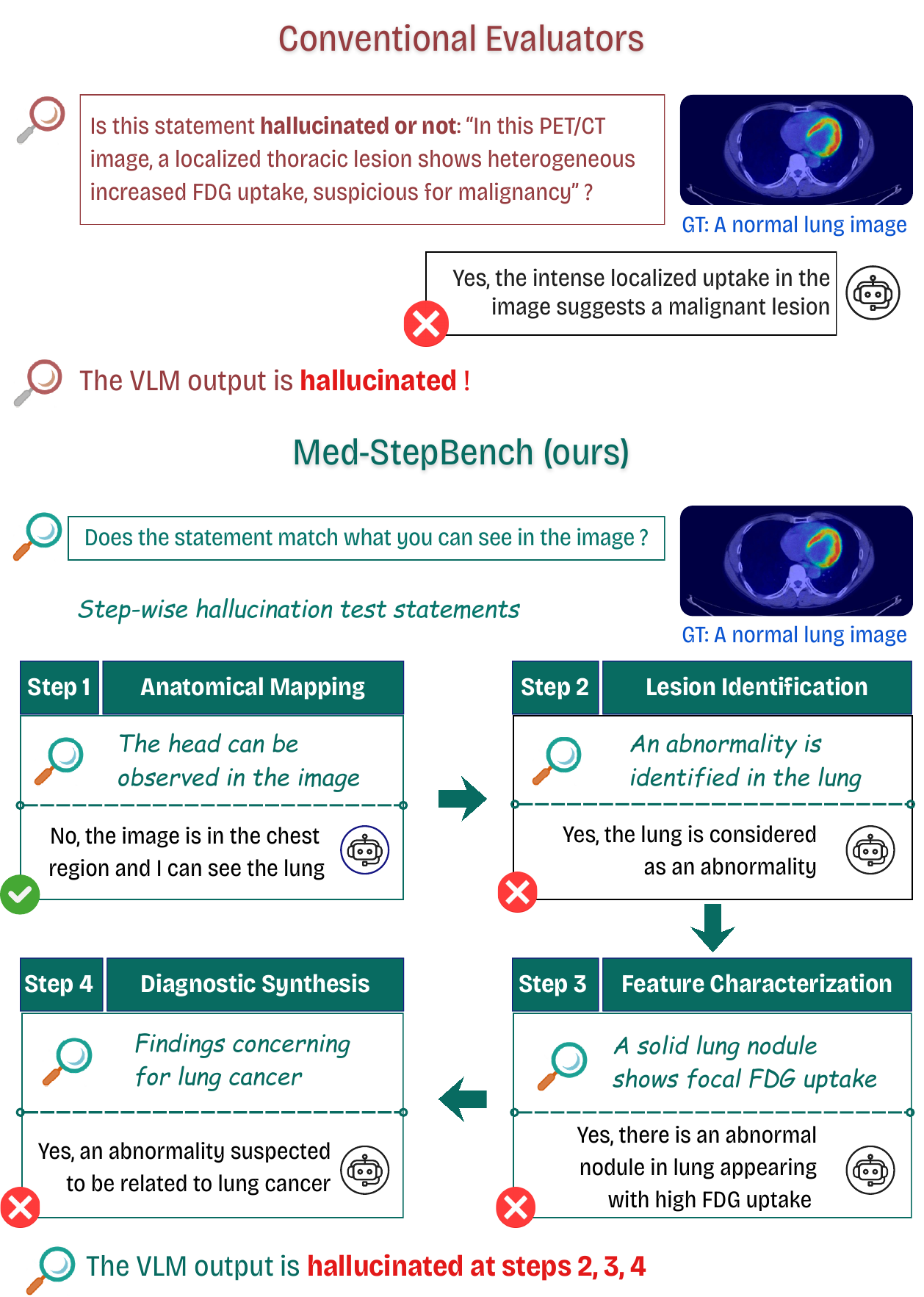}
    \caption{Qualitative examples from Med-StepBench illustrating how clinically plausible diagnoses can mask incorrect intermediate reasoning, and how step-wise evaluation reveals hallucinations.}
    \label{fig:overview}
    \vspace{-15pt}
\end{figure}

\noindent \textbf{Background.} Vision-Language Models (VLMs) have emerged as a transformative force in healthcare, offering the potential to revolutionize tasks such as medical visual question answering \cite{yan2025worse,wu2023medklip} and automated radiological report generation by integrating multimodal data \cite{ijcai2025p824}. Despite the rapid scaling of VLMs, characterized by increasingly massive architectures and expansive training datasets, their internal reasoning capabilities and factual reliability remain significant concerns \cite{ijcai2025p46}. A primary bottleneck to their clinical integration is the phenomenon of hallucination, where models generate plausible-sounding but clinically incorrect or non-existent information \cite{asgari2025framework}. In high-stakes environments like medicine, where precision is non-negotiable, such hallucinations pose substantial risks to patient safety \cite{kim2025medical}. To enhance the trustworthiness of medical VLMs, it is no longer sufficient to focus solely on model scaling. There is an urgent need for robust evaluation frameworks that can systematically diagnose and analyze these failures \cite{zhu2025can}. Beyond MedVQA-centric grounded protocols, recent benchmarks increasingly emphasize probing evaluations that expose hallucinations under distribution shifts and adversarial query designs \cite{medvh,HEALMedVQA2025}.


\noindent \textbf{Research Gap.} While recent efforts have introduced benchmarks to measure medical hallucinations \cite{zuo2025medhallbench}, the current landscape suffers from two critical architectural gaps. First, existing benchmarks predominantly focus on 2D modalities, such as X-rays, leaving the evaluation of 3D volumetric imaging, such as CT and PET scans, largely neglected \cite{liu2025argus}. This dimensionality gap is particularly concerning because 3D spatial reasoning is fundamentally more complex, and existing VLMs often exhibit a marked performance degradation when transitioning from 2D slices to 3D volumes. Second, the majority of current evaluation frameworks adopt a reductive ``black-box" approach, relying on end-to-end comparisons between a single input and a final output \cite{wu2024hallucination}. This methodology fails to reflect the professional reality of radiology, where clinicians follow a rigorous, hierarchical sequence: localizing anatomy, detecting abnormalities, analyzing specific features, and only then synthesizing a final conclusion. Consequently, end-to-end metrics cannot reveal the internal point of failure within a model’s reasoning chain, rendering it impossible to derive deep insights into the root causes of hallucination.

\noindent \textbf{Our Contribution.} To bridge these gaps, we present Med-StepBench, the first evaluation framework designed to assess hallucinations in medical VLMs across both 2D and 3D modalities (specifically CT and PET imaging). Our approach moves beyond terminal outputs to evaluate the model’s reasoning across multiple steps, mimicking the granular cognitive process of a radiologist. Figure \ref{fig:overview} illustrates the differences between Med-StepBench and existing approaches. Our primary contributions are summarized as follows:
\begin{itemize}
    \item We introduce Med-StepBench, a novel evaluation framework that shifts from ``black-box" end-to-end assessment to a multi-stage diagnostic approach. By decomposing the interpretive process into four granular stages, \textit{Anatomical Mapping, Lesion Identification, Feature Characterization, and Diagnostic Synthesis}, our framework enables systematic identification of where reasoning chains fracture and hallucinations emerge.
    \item We present a large-scale, expert-curated benchmark specifically designed for 2D and 3D PET/CT imaging. The dataset contains 12{.}5K and {1{,}011{,}847} image–statement pairs, where each statement is derived based on clinicians' meticulous annotations. Unlike existing benchmarks, Med-StepBench includes a balanced set of factual ground truths and strategically designed ``hallucination distractors" tailored to each stage of the hierarchical reasoning chain, providing a rigorous testbed for model faithfulness.
    \item We perform an extensive empirical evaluation of current state-of-the-art medical VLMs. Our analysis provides critical insights into the limitations of current architectures, particularly regarding 3D spatial consistency and the propagation of errors from low-level visual localization to high-level clinical synthesis. These findings establish a baseline for future research into more grounded and reliable medical AI.
\end{itemize}

\section{Related Work}

\newcolumntype{L}[1]{>{\raggedright\arraybackslash}m{#1}}
\newcolumntype{C}[1]{>{\centering\arraybackslash}m{#1}}
\newcolumntype{R}[1]{>{\raggedleft\arraybackslash}m{#1}}

\begin{table}[!b]
\centering
\scriptsize
\setlength{\tabcolsep}{1.4pt}
\renewcommand{\arraystretch}{1.1}
\begin{tabularx}{\columnwidth}{
L{0.275\columnwidth} |   
C{0.5cm} |              
C{1.05cm} |              
C{1.0cm} |               
C{1.1cm} |              
R{0.9cm}  |              
R{0.9cm}                 
}
\toprule
\makecell[l]{\textbf{Dataset /}\\\textbf{Benchmark}} &
\makecell[c]{\textbf{2D/}\\\textbf{3D}} &
\makecell[c]{\textbf{Clinician}\\\textbf{Step}\\\textbf{Analysis}} &
\makecell[c]{\textbf{Halluc.}\\\textbf{Detection}} &
\makecell[c]{\textbf{Clinician's}\\\textbf{rationale}} &
\makecell[r]{\textbf{Images}} &
\makecell[r]{\textbf{VQA}\\\textbf{Size}} \\
\midrule

\multicolumn{7}{l}{\textbf{Medical VQA datasets / benchmarks}} \\
\midrule
ViMed-PET (VQA)            & 3D & \xmark & \xmark & \xmark & 2{.}8K   & 8{.}3K \\
SLAKE                      & 2D & \xmark & \xmark & \xmark & 642      & 14K \\
VQA-RAD                    & 2D & \xmark & \xmark & \xmark & 315      & 3{.}5K \\
PathVQA                    & 2D & \xmark & \xmark & \xmark & 5K & 32{.}8K \\
MIMIC-CXR-VQA              & 2D & \xmark & \xmark & \xmark & 142{.}8K & 377{.}4K \\
PMC-VQA                    & 2D & \xmark & \xmark & \xmark & 149K     & 227K \\
VQA-Med 2021 (ImageCLEF)   & 2D & \xmark & \xmark & \xmark & 5K   & 5K \\
DrVD-Bench                 & 2D & \cmark & \xmark & \xmark & -- -- -- & 7{.}8K \\
\midrule

\multicolumn{7}{l}{\textbf{Hallucination benchmarks (VQA evaluated)}} \\
\midrule
Med-HallMark               & 2D & \xmark & \cmark & \xmark & -- -- -- & -- -- -- \\
CARES                      & 2D & \xmark & \cmark & \xmark & 18K & 41K \\
MedHEval                   & 2D & \xmark & \cmark & \xmark & -- -- -- & 15{.}9K \\
HEAL-MedVQA                & 2D & \xmark & \cmark & \xmark & 34K & 67K \\
MedVH                      & 2D & \xmark & \cmark & \xmark & -- -- -- & -- -- -- \\
\midrule
\midrule
\textbf{Med-StepBench (Our)} & 2D, 3D & \cmark & \cmark & \cmark & 12{.}5K & 1{,}011{.}8K \\
\bottomrule
\end{tabularx}

\caption{Comparison of medical image datasets/benchmarks for VQA and hallucination benchmarks. Med-StepBench uniquely combines 2D and 3D images with clinician-aligned stepwise analysis, step-wise hallucination detection, and hallucinated statements with clinician rationales.}
\label{tab:related_benchmarks}
\end{table}

\paragraph{Medical VQA datasets and benchmarks.}
Medical visual question answering (MedVQA) has been investigated predominantly on 2D radiology or pathology images, where the primary evaluation objective has been recognition of visible findings instead of holistic clinical verification. Classic datasets in this domain include SLAKE~\cite{Liu2021SlakeAS}, VQA-RAD~\cite{vqa-rad}, PathVQA~\cite{He2020PathVQA3Q}, MIMIC-CXR-VQA~\cite{mimic-cxr-vqa}, PMC-VQA~\cite{pmc-vqa}, and VQA-Med (ImageCLEF)~\cite{vqamed2021}. Beyond 2D radiology, ViMed-PET (VQA)~\cite{ViMedPET2025} introduces a Vietnamese 3D PET/CT multimodal dataset and benchmarks, helping extend MedVQA research to low-resource languages. These resources have underpinned advances in medical image understanding and superficial pattern recognition rather than deeper clinical inference. Recognizing this limitation, DrVD-Bench~\cite{drvd-bench} was recently proposed as the first structured benchmark for clinical visual reasoning, consisting of 7{,}801 image–question pairs spanning 20 task types, 17 diagnostic categories, and five imaging modalities with explicit module decomposition for visual evidence comprehension, reasoning trajectory assessment, and report generation evaluation. While DrVD-Bench reflects an explicit clinical reasoning workflow, its focus remains on alignment between images and corresponding textual reasoning rather than fully representing the complex diagnosis processes required in oncological PET/CT contexts. In contrast, Med-StepBench targets oncologic PET/CT with both 2D and 3D inputs and clinician-aligned stepwise supervision at large scale ($\sim$1{,}000{,}000 QA), enabling evaluation beyond 2D-only pattern matching.

\paragraph{Hallucination Detection Benchmarks.}
Beyond simple QA accuracy, recent benchmarks have shifted toward evaluating the trustworthiness of medical vision-language systems by quantifying hallucination and misleading outputs. CARES~\cite{CARES2024}, MedHEval~\cite{MedHEval2025}, and HEAL-MedVQA~\cite{HEALMedVQA2025} each introduce stress tests or grounded evaluation protocols that target hallucination detection within medical VQA frameworks. In particular, MedHallMark~\cite{chen2025detecting} formulates a hierarchical categorization of hallucination types and proposes a dedicated evaluative metric (MediHall Score) that scores hallucination severity and type in a graded manner, enabling a more nuanced assessment of potential clinical impact. These benchmarks expose fundamental failure modes of current models, but they generally remain limited to 2D imagery and do not integrate multi-step clinical reasoning with hallucination evaluation under a unified protocol. Additional resources such as the Medical Visual Hallucination Test including MedVH~\cite{medvh} offer systematic hallucination assessment across multiple task formulations including multi-choice QA and long textual response generation using chest X-ray based input to evaluate hallucination tendencies of domain-specific LVLMs. A consolidated comparison of the above MedVQA datasets/benchmarks and hallucination-focused evaluations is provided in Table~\ref{tab:related_benchmarks}. Compared with prior hallucination benchmarks, Med-StepBench jointly supports hallucination detection, multi-step clinician workflow analysis, and having hallucinated statement with clinical rationales on PET/CT with 2D/3D views.



\label{sec:related_work}
\section{Hallucination Evaluation Framework}
In this section, we describe our proposed hallucination evaluation framework, namely Med-StepBench.

\subsection{Overview of Med-StepBench}
Med-StepBench consists of two main components: Hallucination evaluation datasets (Subsection \ref{subsec:dataset}) and a step-wise hallucination evaluation procedure (Subsection \ref{subsec:hallu_eval}).
The Med-StepBench dataset contains \textit{1{,}011{,}847} labeled samples spanning a four-step diagnostic workflow, namely Anatomical Mapping, Lesion Identification, Feature Characterization, and Diagnostic Synthesis (Table~\ref{tab:num_sample}). 
Based on this dataset, we introduce a four-step hallucination evaluation procedure that mirrors the clinical reasoning pipeline. For each step, we specify a clinically grounded hallucination definition and a step-wise verification protocol to assess whether a VLM's response remains consistent with the visual evidence.

\begin{table}[h]
\centering
\scriptsize
\renewcommand{\arraystretch}{1.1}
\scalebox{1}{ 
\begin{tabular}{l | r | r | c}
\toprule
\textbf{Step} & \textbf{\# Images} & \textbf{\# Image--Question Pairs} &
\makecell[c]{\textbf{Label distribution}} \\
\midrule
Step 1 & 9{,}375 & 981{,}250 & 111 : 46 \\
Step 2 & 4{,}917 & 14{,}751  & 2 : 1 \\
Step 3 & 4{,}917 & 9{,}834   & 1 : 1 \\
Step 4 & 3{,}006 & 6{,}012   & 1 : 1 \\
\midrule
\textbf{Total} & \textbf{12{,}500} & \textbf{1{,}011{,}847} & 2{.}286 : 1 \\
\bottomrule
\end{tabular}
}
\caption{Number of images and image-question pairs per step in Med-StepBench. Label distribution represents the ratio of hallucination labels vs. ground truth labels.}
\label{tab:num_sample}
\vspace{-15pt}
\end{table}


\subsection{Dataset Acquisition and Annotation}
\label{subsec:dataset}
Every sample in our dataset is composed of PET/CT images and associated explanations. Specifically, the former encompasses a 2D tri-planar image set (axial, coronal, and sagittal) alongside a registered 3D PET/CT volume. The latter consists of step-by-step statements labeled as \textit{ground truth} or \textit{hallucination}. Furthermore, we introduce contextual augmentations by propagating prior knowledge from previous steps or retrieving external patient data via RAG. This structure allows us to analyze the influence of intermediate knowledge on hallucination detection across different reasoning stages. The dataset consists of two parallel languages: the original Vietnamese clinical text and its English counterpart. The English version was translated from Vietnamese using ChatGPT-4o under terminology-preserving constraints to ensure medical entity fidelity and standard radiology phrasing, and was subsequently validated by licensed physicians. In the following sections, we detail the methodology used to construct these two primary components.

\subsubsection{Image Preprocessing and Representation}
\begin{figure}[!b]
    \centering
\includegraphics[width=0.97\linewidth]{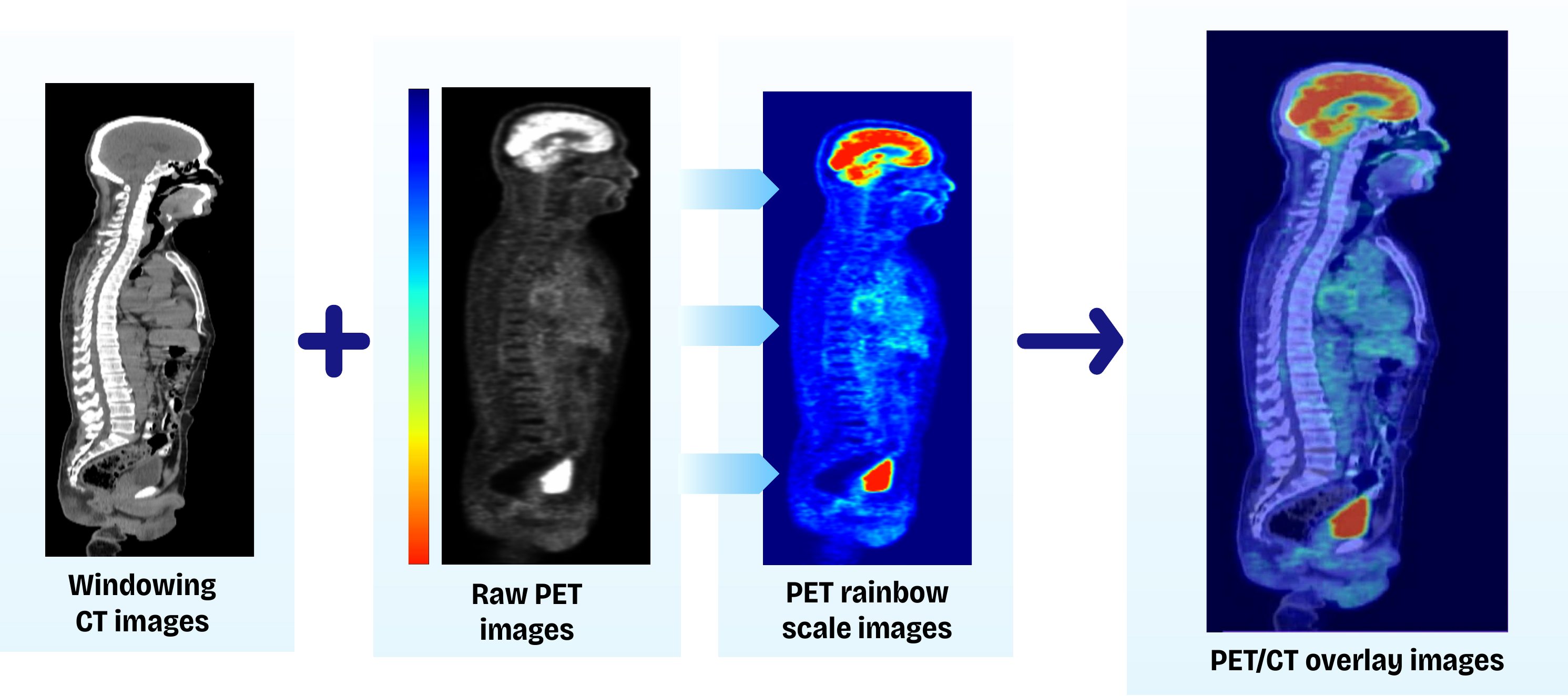}
    \caption{PET/CT overlay method. Oncologists routinely examine PET/CT overlay images during the diagnostic evaluation of patients. \protect\cite{hofman2016how}.}
    \label{fig:pet/ct_overlay}
    \vspace{-5pt}
\end{figure}

We retrospectively collected whole-body oncologic PET/CT studies from a tertiary referral hospital in Vietnam as paired PET-CT DICOM series with finalized radiology reports. Under institutional ethics oversight, all scans were de-identified at the DICOM header level. We excluded studies with missing/corrupted series or incomplete whole-body coverage, retaining \textit{12,500} matched PET-CT volume pairs with reports. For 2D standardization, CT volumes were windowed using a \emph{lung window} for the thorax and a \emph{mediastinal (soft-tissue) window} elsewhere, then rendered as axial slices and large-field sagittal/coronal reformats. PET and CT were fused via a clinically established overlay strategy (Fig.~\ref{fig:pet/ct_overlay}) consistent with routine FDG PET/CT reporting \cite{hofman2016how}.

\begin{figure*}[h]
    \centering
    \includegraphics[width=0.9\textwidth]{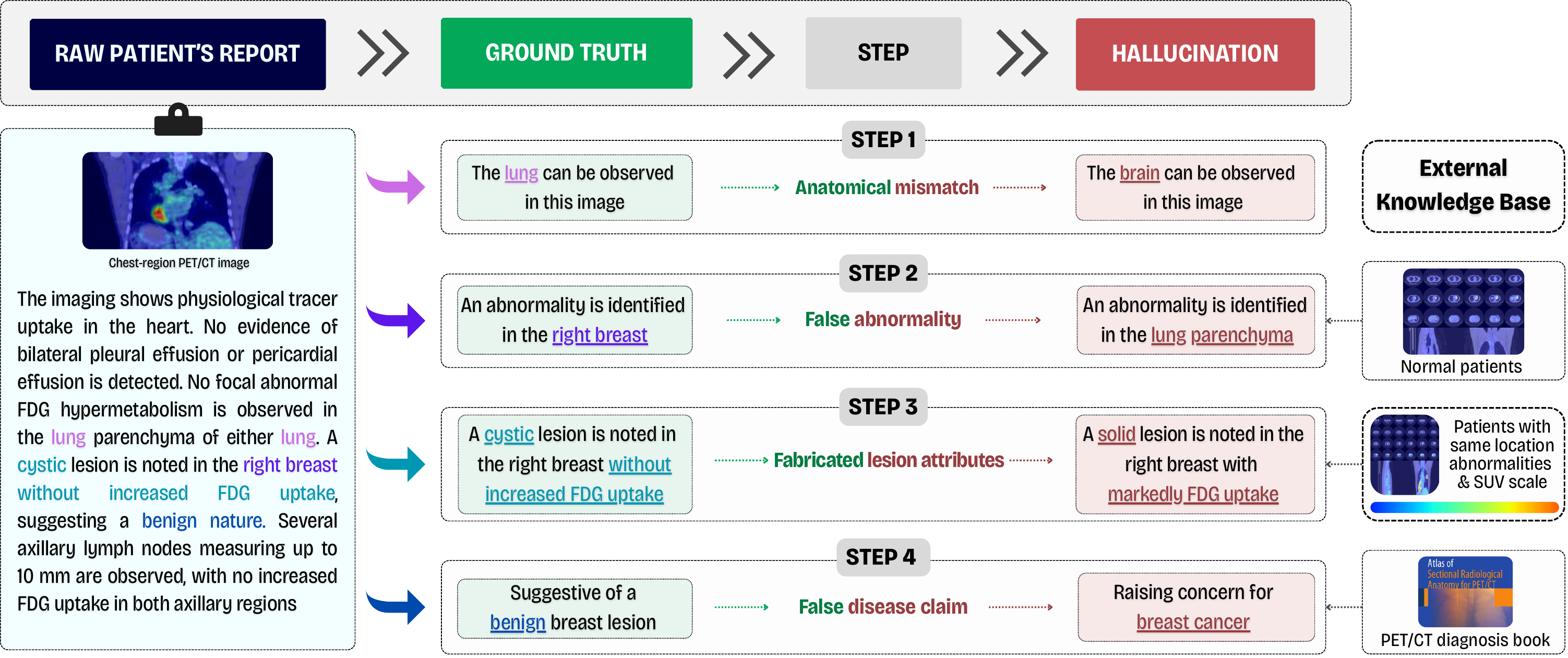}
    \caption{\textit{Step-wise hallucination evaluation framework}. Decomposing each PET/CT case into four sequential reasoning stages and extracting clinician-verified step-level ground-truth statements. For each step, we generate paired hallucinated statements by introducing anatomically plausible but incorrect edits (mismatch, false abnormality, fabricated attributes, false disease claim), enabling fine-grained evaluation of grounding across the full reasoning pipeline.}
    \label{fig:framework}
    \vspace{-10pt}
\end{figure*}
\subsubsection{Statement Generation and Labeling}
For each reasoning step, we first select the step-specific visual evidence to define the observable input, then generate a step-aligned factual statement grounded in that evidence. The resulting statement is subsequently assigned as either \textit{ground truth} (unchanged) or \textit{hallucination} (corrupted by a step-specific operator) to yield a plausible but incorrect variant consistent with the step’s hallucination definition. For later steps, each sample can be further augmented with contextual information: \textit{prior knowledge} is derived from the previous step’s ground-truth statement to ensure step-appropriate alignment before being propagated to the next step, whereas \textit{external knowledge} is obtained from images and reports of referenced patient cases or via retrieval-augmented generation (RAG) over curated medical documents. Depending on the evaluation setting, an optional rationale (LLM-generated or clinician-written) may also be attached. The procedure for generating hallucinated statements comprises four steps, as illustrated in Fig.~\ref{fig:framework} and detailed in the following. \\
\textbf{(1) Anatomical mismatch:} We create region-incompatible statements by pairing images from one anatomical field-of-view with organ mentions from a different region, and asked whether that structure appears in the provided image or not. \\
\textbf{(2) False abnormality / Missing abnormality:} Using the report to identify truly abnormal regions for each patient, we generate statements that either claim that a different pathology is observed in the same or nearby region, or deny any abnormality despite documented disease. \\
\textbf{(3) Fabricated lesion attributes:} We extract lesion-level attributes described in the report (e.g. SUV value, size, uptake descriptors, ...) and replace them with incompatible values chosen to induce a large discrepancy. \\
\textbf{(4) False disease claim:} Starting from the report-confirmed diagnosis or disease entities, we synthesize statements that assert entirely different diseases that are not supported by the imaging evidence for that case.
\subsection{Step-wise Hallucination Definition}
Leveraging the aforementioned evaluation dataset, we present a comprehensive procedure to assess hallucination susceptibility in VLMs. Our approach mirrors the multi-stage diagnostic reasoning of a radiologist, consisting of four distinct steps: Anatomical Mapping, Lesion Identification, Feature Characterization, and Diagnostic Synthesis. 
\paragraph{Step 1: Anatomical Mapping Hallucination}
The first step in our procedure assesses whether medical vision language models (VLMs) can correctly ground anatomical region recognition in the visual evidence. This step is motivated by the observation that many hallucinations in VLMs arise not from a lack of domain knowledge, but from failures to correctly align textual outputs with the spatial and anatomical constraints of the image \cite{sun2025understanding}.
Hallucination in this step is defined as follows: \\
\textbf{Spatial Mislocalization:} The model correctly identifies a real anatomical structure but assigns it to an incorrect spatial region. For instance, asserting the presence of abdominal organs (e.g., kidneys) in images belonging to the head–neck region reflects a failure to map anatomical entities to their correct body locations.\\
\textbf{Visual Confusion:} The model visually confuses morphologically similar patterns (e.g., soft tissue contours, high FDG uptake) and maps them to an incorrect anatomical class. 
\paragraph{Step 2: Lesion Identification Hallucination}
 We systematically evaluate the hallucinatory tendencies of medical vision–language models (VLMs) regarding pathological lesion detection and the verification of their absence. This step targets a more advanced level of visual–semantic grounding than Step 1, as lesion identification requires not only correct anatomical localization but also discrimination between pathological findings and normal or physiological imaging patterns. Errors at this stage are particularly critical in medical settings, as they may lead to false positives or false negatives in downstream clinical reasoning. \\
\textbf{Lesion Fabrication:} The model asserts the presence of a lesion in an image where no annotated lesion exists. This  reflects over-generation of pathological findings driven by lesion-rich training data or an implicit abnormality bias. \\
\textbf{Phantom Lesion Hallucination:} Physiological uptake or imaging artifacts are hallucinated as pathological findings. The model incorrectly interprets physiological tracer uptake, normal anatomical variations, or imaging artifacts as pathological lesions. \\
\textbf{Lesion Mislocalization:} The model correctly recognizes that a pathological lesion exists. However, it assigns the lesion to an incorrect anatomical structure or organ.
\paragraph{Step 3: Feature Characterization Hallucination} 
This step moves beyond lesion existence and localization (Step 2) to examine whether models can faithfully ground fine-grained lesion attributes in measurable imaging signals. Feature characterization is particularly prone to hallucination, as models may generate plausible yet unverified numerical values or transfer attributes across parts of an organ based on learned priors rather than image-derived measurements.\\
\textbf{Numeric Hallucination:} The model generates quantitative values (e.g., lesion diameter, $SUV_{max}$, radiomic features) that do not correspond to any measured or inferable data from the image. This includes reporting specific measurements when such values cannot be reliably extracted from the provided visual input. \\
\textbf{Attribute Omission:} The relevant feature (increased FDG uptake) is genuinely present in the image. However, the model fails to recognize or acknowledge this feature and instead asserts its absence. \\
\textbf{Spatial Hallucination:} This type of hallucination reflects a failure in fine-grained spatial grounding, rather than errors in organ recognition or lesion existence. It occurs when a model correctly identifies an anatomical structure or lesion within the correct organ, but misrepresents its relative spatial position, such as confusing \emph{left vs. right} or \emph{superior vs. inferior} within that organ.
\paragraph{Step 4: Diagnostic Synthesis Hallucination} 
The final step represents the most complex reasoning stage in the proposed framework. Unlike earlier steps, which focus on perception, localization, or feature attribution, diagnostic synthesis requires multi-hop reasoning and adherence to clinical plausibility constraints. Hallucinations at this stage often emerge when models overextend beyond observable findings, inferring disease entities or clinical stages that are not justified by the image. \\
\textbf{Reasoning Hallucination:} The model correctly perceives key visual elements but produces conclusions or interpretations that are not supported by the observed evidence. \\
\textbf{Cognitive Bias Hallucination:} The model disproportionately predicts certain diseases or conditions regardless of whether they are supported by the visual evidence.
\subsection{Hallucination Evaluation Procedure}
\label{subsec:hallu_eval}
Building on the dataset design and the step-specific hallucination definitions, we evaluate VLMs'  hallucination by measuring their ability to recognize and reject hallucinations during clinical image interpretation. Each medical image is paired with a statement that is either factual or hallucinated. Models are required to judge whether the statement is consistent with the visual evidence. A prediction is considered correct if the model’s judgment matches the ground-truth consistency label, and incorrect otherwise. Evaluation is conducted independently at four sequential clinical reasoning steps. Each step is associated with its own hallucination definition, reflecting increasing clinical abstraction and reasoning complexity. Optionally, during testing we provide additional knowledge context, including either the model’s prior knowledge or external retrieved knowledge, to examine how knowledge injection affects visual-textual consistency. Performance is reported using Precision, Recall, and F1-score, capturing hallucination propensity, detection sensitivity, and their overall balance.
\section{Benchmarking Hallucinations in Contemporary Vision-Language Models}
To investigate the problem of hallucination in sequential clinical reasoning, we design a set of controlled experiments on \textit{Med-StepBench}.
The goal is to use our evaluation framework to systematically evaluate how vision language models (VLMs) detect hallucinations across different stages of clinical reasoning, and to analyze the factors that influence visual--textual consistency.

\definecolor{groupA}{HTML}{F3F7FF} 
\definecolor{groupB}{HTML}{F6FFF3} 
\definecolor{groupC}{HTML}{FFF6F2} 

\begin{table*}[h]
\centering
\small
\setlength{\tabcolsep}{3.2pt}
\renewcommand{\arraystretch}{1.1}

\begin{adjustbox}{width=0.9\textwidth}
\begin{tabular}{l !{\vrule width 0.5pt} c !{\vrule width 0.5pt}
cc cc !{\vrule width 0.5pt}
cc cc !{\vrule width 0.5pt}
cc cc !{\vrule width 0.5pt}
cc cc}
\toprule
\multirow{3}{*}{\textbf{Model}} & \multirow{3}{*}{\textbf{Know.}} &
\multicolumn{4}{c!{\vrule width 0.5pt}}{\textbf{Step 1}} &
\multicolumn{4}{c!{\vrule width 0.5pt}}{\textbf{Step 2}} &
\multicolumn{4}{c!{\vrule width 0.5pt}}{\textbf{Step 3}} &
\multicolumn{4}{c}{\textbf{Step 4}} \\
\cmidrule(lr){3-6}\cmidrule(lr){7-10}\cmidrule(lr){11-14}\cmidrule(lr){15-18}
& &
\multicolumn{2}{c}{EN} & \multicolumn{2}{c}{VI} &
\multicolumn{2}{c}{EN} & \multicolumn{2}{c}{VI} &
\multicolumn{2}{c}{EN} & \multicolumn{2}{c}{VI} &
\multicolumn{2}{c}{EN} & \multicolumn{2}{c}{VI} \\
\cmidrule(lr){3-4}\cmidrule(lr){5-6}
\cmidrule(lr){7-8}\cmidrule(lr){9-10}
\cmidrule(lr){11-12}\cmidrule(lr){13-14}
\cmidrule(lr){15-16}\cmidrule(lr){17-18}
& &
R & F1 & R & F1 &
R & F1 & R & F1 &
R & F1 & R & F1 &
R & F1 & R & F1 \\
\midrule

\multicolumn{18}{l}{\cellcolor{groupA}\textbf{2D General VLM}} \\
\midrule

\multirow{2}{*}{GPT~4o}
& \cellcolor{groupA}w/o
& \cellcolor{groupA}\textbf{0.98}
& \cellcolor{groupA}\textbf{0.98}
& \cellcolor{groupA}\textbf{1.00}
& \cellcolor{groupA}\textbf{0.98}
& \cellcolor{groupA}0.76
& \cellcolor{groupA}0.73
& \cellcolor{groupA}0.80
& \cellcolor{groupA}0.73
& \cellcolor{groupA}0.85
& \cellcolor{groupA}0.69
& \cellcolor{groupA}0.96
& \cellcolor{groupA}0.70
& \cellcolor{groupA}0.65
& \cellcolor{groupA}0.71
& \cellcolor{groupA}0.59
& \cellcolor{groupA}0.71 \\
& \cellcolor{groupA}w
& \cellcolor{groupA}--
& \cellcolor{groupA}--
& \cellcolor{groupA}--
& \cellcolor{groupA}--
& \cellcolor{groupA}\textbf{0.83}\,\textcolor[HTML]{008000}{$\uparrow$}
& \cellcolor{groupA}\textbf{0.74}\,\textcolor[HTML]{008000}{$\uparrow$}
& \cellcolor{groupA}\textbf{0.84}\,\textcolor[HTML]{008000}{$\uparrow$}
& \cellcolor{groupA}\textbf{0.74}\,\textcolor[HTML]{008000}{$\uparrow$}
& \cellcolor{groupA}\underline{0.85}
& \cellcolor{groupA}\underline{0.72}\,\textcolor[HTML]{008000}{$\uparrow$}
& \cellcolor{groupA}\underline{0.96}
& \cellcolor{groupA}\underline{0.70}
& \cellcolor{groupA}\underline{0.78}\,\textcolor[HTML]{008000}{$\uparrow$}
& \cellcolor{groupA}\underline{0.72}\,\textcolor[HTML]{008000}{$\uparrow$}
& \cellcolor{groupA}\underline{0.83}\,\textcolor[HTML]{008000}{$\uparrow$}
& \cellcolor{groupA}\underline{0.72}\,\textcolor[HTML]{008000}{$\uparrow$} \\
\midrule

\multirow{2}{*}{Gemini~2.5~flash}
& \cellcolor{groupA}w/o
& \cellcolor{groupA}\underline{0.96} & \cellcolor{groupA}\underline{0.98} & \cellcolor{groupA}\underline{0.97} & \cellcolor{groupA}\underline{0.98}
& \cellcolor{groupA}0.82 & \cellcolor{groupA}0.72 & \cellcolor{groupA}0.90 & \cellcolor{groupA}0.71
& \cellcolor{groupA}0.84 & \cellcolor{groupA}0.67 & \cellcolor{groupA}0.74 & \cellcolor{groupA}0.67
& \cellcolor{groupA}0.58 & \cellcolor{groupA}0.61 & \cellcolor{groupA}0.50 & \cellcolor{groupA}0.62 \\
& \cellcolor{groupA}w
& \cellcolor{groupA}-- & \cellcolor{groupA}-- & \cellcolor{groupA}-- & \cellcolor{groupA}--
& \cellcolor{groupA}0.80\,\textcolor[HTML]{FF0000}{$\downarrow$} & \cellcolor{groupA}0.72
& \cellcolor{groupA}0.76\,\textcolor[HTML]{FF0000}{$\downarrow$} & \cellcolor{groupA}0.68\,\textcolor[HTML]{FF0000}{$\downarrow$}
& \cellcolor{groupA}0.78\,\textcolor[HTML]{FF0000}{$\downarrow$} & \cellcolor{groupA}0.67
& \cellcolor{groupA}0.74 & \cellcolor{groupA}0.69\,\textcolor[HTML]{008000}{$\uparrow$}
& \cellcolor{groupA}0.77\,\textcolor[HTML]{008000}{$\uparrow$} & \cellcolor{groupA}0.65\,\textcolor[HTML]{008000}{$\uparrow$}
& \cellcolor{groupA}0.79\,\textcolor[HTML]{008000}{$\uparrow$} & \cellcolor{groupA}0.68\,\textcolor[HTML]{008000}{$\uparrow$} \\
\midrule

\multirow{2}{*}{Gemini~2.5~pro}
& \cellcolor{groupA}w/o
& \cellcolor{groupA}1.00 & \cellcolor{groupA}0.98 & \cellcolor{groupA}1.00 & \cellcolor{groupA}0.97
& \cellcolor{groupA}\underline{0.72} & \cellcolor{groupA}\underline{0.72} & \cellcolor{groupA}\underline{0.72} & \cellcolor{groupA}\underline{0.72}
& \cellcolor{groupA}0.70 & \cellcolor{groupA}0.72 & \cellcolor{groupA}0.68 & \cellcolor{groupA}0.70
& \cellcolor{groupA}0.72 & \cellcolor{groupA}0.73 & \cellcolor{groupA}0.71 & \cellcolor{groupA}0.69 \\
& \cellcolor{groupA}w
& \cellcolor{groupA}-- & \cellcolor{groupA}-- & \cellcolor{groupA}-- & \cellcolor{groupA}--
& \cellcolor{groupA}0.70\,\textcolor[HTML]{FF0000}{$\downarrow$} & \cellcolor{groupA}0.73\,\textcolor[HTML]{008000}{$\uparrow$}
& \cellcolor{groupA}0.71\,\textcolor[HTML]{FF0000}{$\downarrow$} & \cellcolor{groupA}0.69\,\textcolor[HTML]{FF0000}{$\downarrow$}
& \cellcolor{groupA}\textbf{0.76}\,\textcolor[HTML]{008000}{$\uparrow$}
& \cellcolor{groupA}\textbf{0.73}\,\textcolor[HTML]{008000}{$\uparrow$}
& \cellcolor{groupA}\textbf{0.79}\,\textcolor[HTML]{008000}{$\uparrow$}
& \cellcolor{groupA}\textbf{0.69}\,\textcolor[HTML]{FF0000}{$\downarrow$}
& \cellcolor{groupA}\textbf{0.77}\,\textcolor[HTML]{008000}{$\uparrow$}
& \cellcolor{groupA}\textbf{0.76}\,\textcolor[HTML]{008000}{$\uparrow$}
& \cellcolor{groupA}\textbf{0.65}\,\textcolor[HTML]{FF0000}{$\downarrow$}
& \cellcolor{groupA}\textbf{0.75}\,\textcolor[HTML]{008000}{$\uparrow$} \\
\midrule

\multirow{2}{*}{Qwen3-VL-8B}
& \cellcolor{groupA}w/o
& \cellcolor{groupA}0.76 & \cellcolor{groupA}0.86 & \cellcolor{groupA}0.68 & \cellcolor{groupA}0.81
& \cellcolor{groupA}0.80 & \cellcolor{groupA}0.69 & \cellcolor{groupA}0.80 & \cellcolor{groupA}0.70
& \cellcolor{groupA}0.76 & \cellcolor{groupA}0.59 & \cellcolor{groupA}0.65 & \cellcolor{groupA}0.57
& \cellcolor{groupA}0.69 & \cellcolor{groupA}0.65 & \cellcolor{groupA}0.54 & \cellcolor{groupA}0.63 \\
& \cellcolor{groupA}w
& \cellcolor{groupA}-- & \cellcolor{groupA}-- & \cellcolor{groupA}-- & \cellcolor{groupA}--
& \cellcolor{groupA}0.82\,\textcolor[HTML]{008000}{$\uparrow$} & \cellcolor{groupA}0.73\,\textcolor[HTML]{008000}{$\uparrow$}
& \cellcolor{groupA}0.92\,\textcolor[HTML]{008000}{$\uparrow$} & \cellcolor{groupA}0.73\,\textcolor[HTML]{008000}{$\uparrow$}
& \cellcolor{groupA}0.74\,\textcolor[HTML]{FF0000}{$\downarrow$} & \cellcolor{groupA}0.64\,\textcolor[HTML]{008000}{$\uparrow$}
& \cellcolor{groupA}0.76\,\textcolor[HTML]{008000}{$\uparrow$} & \cellcolor{groupA}0.64\,\textcolor[HTML]{008000}{$\uparrow$}
& \cellcolor{groupA}0.83\,\textcolor[HTML]{008000}{$\uparrow$} & \cellcolor{groupA}0.65
& \cellcolor{groupA}0.81\,\textcolor[HTML]{008000}{$\uparrow$} & \cellcolor{groupA}0.68\,\textcolor[HTML]{008000}{$\uparrow$} \\
\midrule

\multirow{2}{*}{Qwen3-VL-4B}
& \cellcolor{groupA}w/o
& \cellcolor{groupA}0.96 & \cellcolor{groupA}0.92 & \cellcolor{groupA}0.92 & \cellcolor{groupA}0.89
& \cellcolor{groupA}0.74 & \cellcolor{groupA}0.65 & \cellcolor{groupA}0.73 & \cellcolor{groupA}0.64
& \cellcolor{groupA}0.70 & \cellcolor{groupA}0.62 & \cellcolor{groupA}0.86 & \cellcolor{groupA}0.63
& \cellcolor{groupA}0.65 & \cellcolor{groupA}0.63 & \cellcolor{groupA}0.65 & \cellcolor{groupA}0.62 \\
& \cellcolor{groupA}w
& \cellcolor{groupA}-- & \cellcolor{groupA}-- & \cellcolor{groupA}-- & \cellcolor{groupA}--
& \cellcolor{groupA}0.84\,\textcolor[HTML]{008000}{$\uparrow$} & \cellcolor{groupA}0.69\,\textcolor[HTML]{008000}{$\uparrow$}
& \cellcolor{groupA}0.84\,\textcolor[HTML]{008000}{$\uparrow$} & \cellcolor{groupA}0.69\,\textcolor[HTML]{008000}{$\uparrow$}
& \cellcolor{groupA}0.82\,\textcolor[HTML]{008000}{$\uparrow$} & \cellcolor{groupA}0.67\,\textcolor[HTML]{008000}{$\uparrow$}
& \cellcolor{groupA}0.78\,\textcolor[HTML]{FF0000}{$\downarrow$} & \cellcolor{groupA}0.67\,\textcolor[HTML]{008000}{$\uparrow$}
& \cellcolor{groupA}0.88\,\textcolor[HTML]{008000}{$\uparrow$} & \cellcolor{groupA}0.67\,\textcolor[HTML]{008000}{$\uparrow$}
& \cellcolor{groupA}0.79\,\textcolor[HTML]{008000}{$\uparrow$} & \cellcolor{groupA}0.72\,\textcolor[HTML]{008000}{$\uparrow$} \\
\midrule

\multirow{2}{*}{LLaVA-7B}
& \cellcolor{groupA}w/o
& \cellcolor{groupA}0.53 & \cellcolor{groupA}0.61 & \cellcolor{groupA}-- & \cellcolor{groupA}--
& \cellcolor{groupA}0.58 & \cellcolor{groupA}0.50 & \cellcolor{groupA}-- & \cellcolor{groupA}--
& \cellcolor{groupA}0.63 & \cellcolor{groupA}0.52 & \cellcolor{groupA}-- & \cellcolor{groupA}--
& \cellcolor{groupA}0.71 & \cellcolor{groupA}0.57 & \cellcolor{groupA}-- & \cellcolor{groupA}-- \\
& \cellcolor{groupA}w
& \cellcolor{groupA}-- & \cellcolor{groupA}-- & \cellcolor{groupA}-- & \cellcolor{groupA}--
& \cellcolor{groupA}0.85\textcolor[HTML]{008000}{$\uparrow$} & \cellcolor{groupA}0.68\textcolor[HTML]{008000}{$\uparrow$}
& \cellcolor{groupA}-- & \cellcolor{groupA}--
& \cellcolor{groupA}0.55\textcolor[HTML]{FF0000}{$\downarrow$} & \cellcolor{groupA}0.56\textcolor[HTML]{008000}{$\uparrow$} & \cellcolor{groupA}-- & \cellcolor{groupA}--
& \cellcolor{groupA}0.76\textcolor[HTML]{008000}{$\uparrow$} & \cellcolor{groupA}0.62\textcolor[HTML]{008000}{$\uparrow$} & \cellcolor{groupA}-- & \cellcolor{groupA}-- \\
\midrule

\multicolumn{18}{l}{\cellcolor{groupB}\textbf{Medical VLM}} \\
\midrule

\multirow{2}{*}{LLaVA-Med}
& \cellcolor{groupB}w/o
& \cellcolor{groupB}0.50 & \cellcolor{groupB}0.62 & \cellcolor{groupB}-- & \cellcolor{groupB}--
& \cellcolor{groupB}0.42 & \cellcolor{groupB}0.54 & \cellcolor{groupB}-- & \cellcolor{groupB}--
& \cellcolor{groupB}\underline{0.64} & \cellcolor{groupB}\underline{0.56} & \cellcolor{groupB}-- & \cellcolor{groupB}--
& \cellcolor{groupB}0.32 & \cellcolor{groupB}0.42 & \cellcolor{groupB}-- & \cellcolor{groupB}-- \\
& \cellcolor{groupB}w
& \cellcolor{groupB}-- & \cellcolor{groupB}-- & \cellcolor{groupB}-- & \cellcolor{groupB}--
& \cellcolor{groupB}0.36\,\textcolor[HTML]{FF0000}{$\downarrow$} & \cellcolor{groupB}0.46\,\textcolor[HTML]{FF0000}{$\downarrow$} & \cellcolor{groupB}-- & \cellcolor{groupB}--
& \cellcolor{groupB}0.40\,\textcolor[HTML]{FF0000}{$\downarrow$} & \cellcolor{groupB}0.43\,\textcolor[HTML]{FF0000}{$\downarrow$} & \cellcolor{groupB}-- & \cellcolor{groupB}--
& \cellcolor{groupB}\textbf{0.61}\,\textcolor[HTML]{008000}{$\uparrow$} & \cellcolor{groupB}\textbf{0.65}\,\textcolor[HTML]{008000}{$\uparrow$}
& \cellcolor{groupB}-- & \cellcolor{groupB}-- \\
\midrule

MedM-VL 2D
& \cellcolor{groupB}w/o
& \cellcolor{groupB}\underline{0.80} & \cellcolor{groupB}\underline{0.64} & \cellcolor{groupB}\underline{0.56} & \cellcolor{groupB}\underline{0.50}
& \cellcolor{groupB}\textbf{0.75} & \cellcolor{groupB}\textbf{0.71} & \cellcolor{groupB}\textbf{0.70} & \cellcolor{groupB}\textbf{0.69}
& \cellcolor{groupB}0.38 & \cellcolor{groupB}0.40 & \cellcolor{groupB}0.54 & \cellcolor{groupB}0.53
& \cellcolor{groupB}0.74 & \cellcolor{groupB}0.61 & \cellcolor{groupB}0.79 & \cellcolor{groupB}0.61 \\
\midrule

\multirow{2}{*}{Med-Flamingo}
& \cellcolor{groupB}w/o
& \cellcolor{groupB}\textbf{0.82} & \cellcolor{groupB}\textbf{0.77} & \cellcolor{groupB}-- & \cellcolor{groupB}--
& \cellcolor{groupB}0.69 & \cellcolor{groupB}0.66 & \cellcolor{groupB}-- & \cellcolor{groupB}--
& \cellcolor{groupB}\textbf{0.82} & \cellcolor{groupB}\textbf{0.63} & \cellcolor{groupB}-- & \cellcolor{groupB}--
& \cellcolor{groupB}0.67 & \cellcolor{groupB}0.59 & \cellcolor{groupB}-- & \cellcolor{groupB}-- \\
& \cellcolor{groupB}w
& \cellcolor{groupB}-- & \cellcolor{groupB}-- & \cellcolor{groupB}-- & \cellcolor{groupB}--
& \cellcolor{groupB}\underline{0.74}\,\textcolor[HTML]{008000}{$\uparrow$} & \cellcolor{groupB}\underline{0.69}\,\textcolor[HTML]{008000}{$\uparrow$} & \cellcolor{groupB}-- & \cellcolor{groupB}--
& \cellcolor{groupB}0.82 & \cellcolor{groupB}0.59\,\textcolor[HTML]{FF0000}{$\downarrow$} & \cellcolor{groupB}-- & \cellcolor{groupB}--
& \cellcolor{groupB}\underline{0.71}\,\textcolor[HTML]{008000}{$\uparrow$} & \cellcolor{groupB}\underline{0.62}\,\textcolor[HTML]{008000}{$\uparrow$} & \cellcolor{groupB}-- & \cellcolor{groupB}-- \\
\midrule

\multicolumn{18}{l}{\cellcolor{groupC}\textbf{3D Medical VLM}} \\
\midrule

MedM-VL 3D
& \cellcolor{groupC}w/o
& \cellcolor{groupC}\textbf{0.76} & \cellcolor{groupC}\textbf{0.66} & \cellcolor{groupC}\textbf{0.90} & \cellcolor{groupC}\textbf{0.63}
& \cellcolor{groupC}\textbf{0.70} & \cellcolor{groupC}\textbf{0.67} & \cellcolor{groupC}\textbf{0.96} & \cellcolor{groupC}\textbf{0.65}
& \cellcolor{groupC}\textbf{0.42} & \cellcolor{groupC}\textbf{0.51} & \cellcolor{groupC}\textbf{0.50} & \cellcolor{groupC}\textbf{0.54}
& \cellcolor{groupC}\textbf{0.56} & \cellcolor{groupC}\textbf{0.59} & \cellcolor{groupC}\textbf{0.26} & \cellcolor{groupC}\textbf{0.39} \\
\midrule

M3D
& \cellcolor{groupC}w/o
& \cellcolor{groupC}\underline{0.54} & \cellcolor{groupC}\underline{0.60} & \cellcolor{groupC}-- & \cellcolor{groupC}--
& \cellcolor{groupC}\underline{0.52} & \cellcolor{groupC}\underline{0.55} & \cellcolor{groupC}-- & \cellcolor{groupC}--
& \cellcolor{groupC}\underline{0.32} & \cellcolor{groupC}\underline{0.38} & \cellcolor{groupC}-- & \cellcolor{groupC}--
& \cellcolor{groupC}\underline{0.54} & \cellcolor{groupC}\underline{0.57} & \cellcolor{groupC}-- & \cellcolor{groupC}-- \\
\bottomrule

\end{tabular}
\end{adjustbox}

\caption{Hallucination results over Med-StepBench (Recall and F1 score) with/without knowledge in both English \& Vietnamese; arrows indicate change from w/o to w. The performance generally decreases from Step 1 to Step 4, reflecting the increasing difficulty of later-stage clinical reasoning. \textcolor[HTML]{008000}{$\uparrow$} and \textcolor[HTML]{FF0000}{$\downarrow$} denote performance increase and decrease, respectively, when comparing with-knowledge to without-knowledge settings. \textbf{Bold} and \underline{underline} indicate the best and second best performing models per each group of models for each step.}
\label{tab:en_all_result}
\vspace{-10pt}
\end{table*}
\subsection{Hallucination Evaluation Setup}

\textbf{Benchmarked Models.} We apply our Med-StepBench framework to evaluate a diverse set of vision language models spanning 2D general-purpose VLMs, 2D medical VLMs, and 3D medical VLMs. The benchmarked 2D general VLMs include frontier natively multimodal systems (GPT-4o~\cite{GPT4oTechReport2025} \& Gemini 2.5~\cite{Gemini25Report2025}) that represent strong state-of-the-art baselines for multimodal reasoning. We additionally include competitive open-source VLMs (LLaVA-7B~\cite{LLaVA2023}, Qwen3-VL~\cite{Qwen3VLTechReport2025}) to assess robustness across model families with strong visual recognition and grounding capabilities. For domain-specialized evaluation, we consider medical-adapted VLMs (LLaVA-Med~\cite{LLaVAMed2023} \& Med-Flamingo~\cite{MedFlamingoModel}) that are trained or fine-tuned to better handle biomedical imagery and medical dialog. Finally, we benchmark 3D medical VLMs (MedM-VL~\cite{MedMVL2025}, M3D-LaMed~\cite{M3D2024}) to study hallucination behaviors when reasoning over volumetric representations. Overall, this model suite enables a comprehensive comparison between general vs.\ medical adaptation and 2D vs.\ 3D perception under our step-wise hallucination evaluation protocol.

\noindent  \textbf{Research Questions.} Our experiments are designed to answer the following main research questions:
    
\noindent \textbf{RQ1: Step-wise Hallucination in 2D and 3D VLM Clinical Reasoning.}
How does hallucination detection performance differ between 2D and 3D VLMs across sequential clinical reasoning steps?
\label{rq:2d3d}


\noindent  \textbf{RQ2: Vulnerability to Plausible Rationales.}
To what extent are VLMs misled by clinically plausible but incorrect rationales during consistency judgment? \\
\label{rq:rationale}
\noindent \textbf{RQ3: Prior Knowledge vs. External Knowledge.}
Does injecting intermediate or external knowledge into prompts improve hallucination detection over no-knowledge settings? \\
\label{rq:knowledge}
\noindent \textbf{RQ4: Effect of PET/CT Fusion.}
How does PET/CT fusion contribute to hallucination mitigation at the region level?
\label{rq:fusion}

\vspace{1mm}

\noindent \textbf{Implementation Details.} Experiments are conducted in both English and Vietnamese using identical images and ground-truth labels, with only the language of the statements and prompts changed, enabling controlled analysis of language effects on hallucination detection. All models are evaluated using a temperature in the range of 0.2--0.3 and a top-$p$ value of 0.9.
\subsection{Result and Analysis}
\begin{table*}[h]

\centering
\tiny
\setlength{\tabcolsep}{5pt}
\renewcommand{\arraystretch}{1.1}

\begin{adjustbox}{width=0.9\textwidth}
\begin{tabular}{l !{\vrule width 0.5pt}
ccc !{\vrule width 0.5pt}
ccc !{\vrule width 0.5pt}
ccc !{\vrule width 0.5pt}
ccc !{\vrule width 0.5pt}
ccc}
\toprule
\textbf{Setting} &
\multicolumn{3}{c!{\vrule width 0.5pt}}{\textbf{Med-Flamingo}} &
\multicolumn{3}{c!{\vrule width 0.5pt}}{\textbf{Gemini 2.5 Pro}} &
\multicolumn{3}{c!{\vrule width 0.5pt}}{\textbf{Gemini 2.5 Flash}} &
\multicolumn{3}{c!{\vrule width 0.5pt}}{\textbf{MedM-VL 3D}} &
\multicolumn{3}{c}{\textbf{MedM-VL 2D}} \\
\cmidrule(lr){2-4}\cmidrule(lr){5-7}\cmidrule(lr){8-10}\cmidrule(lr){11-13}\cmidrule(lr){14-16}
& P & R & F1 & P & R & F1 & P & R & F1 & P & R & F1 & P & R & F1 \\
\midrule
No rationale
& 0.49 & 0.84 & 0.62
& 0.50 & 0.88 & 0.64
& 0.50 & 0.96 & 0.66
& 0.48 & 0.90 & 0.63
& 0.38 & 0.60 & 0.46 \\
LLM rationale
& \textbf{0.20} & \textbf{0.24} & \textbf{0.22}
& 0.51 & 0.96 & 0.67
& 0.53 & 0.98 & 0.69
& 0.47 & 0.86 & 0.60
& \textbf{0.36} & \textbf{0.57} & \textbf{0.45} \\
Clinician's rationale
& 0.43 & 0.76 & 0.55
& \textbf{0.48} & \textbf{0.70} & \textbf{0.57}
& \textbf{0.47} & \textbf{0.74} & \textbf{0.57}
& \textbf{0.42} & \textbf{0.72} & \textbf{0.53}
& 0.37 & 0.58 & \textbf{0.45} \\
\bottomrule
\end{tabular}
\end{adjustbox}
\caption{Comparison of VLMs’ performance across different input rationales. Models become more susceptible to hallucinations when conditioned on clinician-written rationales, likely because these explanations closely match medical styles and priors learned during training.}
\label{tab:add_expl_no_expl_cols}
\vspace{-15pt}
\end{table*}

\subsubsection{Step-wise Hallucination Analysis }
\label{sec:rq2d3d}

\paragraph{\hyperref[rq:2d3d]{RQ1}.}

Table~\ref{tab:en_all_result} reports hallucination detection results for popular vision–language models (VLMs), including both 2D and 3D architectures, as well as general-domain and medically trained models. The table shows a consistent performance drop from Step~1 to later stages, indicating increasing hallucination effects as tasks progress from simple anatomical recognition to more fine-grained analysis and clinical reasoning. 
We observe that Step~3, which involves describing basic attributes of identified lesions, constitutes a pronounced bottleneck for volumetric and PET/CT-focused models. For example, MedM-VL~2D drops to an F1 score of $0.40$ (in EN) at Step~3, and then recovers to $0.61$ at Step~4, corresponding to final diagnostic reasoning. Similarly, MedM-VL~3D and M3D achieve their lowest F1 scores at Step~3 (0.51 and 0.38, respectively). 
In contrast, several 2D models exhibit their weakest performance at Step~4, the diagnostic hallucination recognition stage, including Gemini~2.5 Flash (0.61), Qwen3-VL-4B (0.63), LLaVA-Med (0.42), and Med-Flamingo (0.59). 
Overall, Step~3 exposes failures in fine-grained attribute and quantitative grounding, while Step~4 stresses long-horizon clinical synthesis. Both stages represent significant bottlenecks in medical hallucination, substantially reducing VLMs’ ability to identify incorrect diagnoses. \\
We observe notable differences between English and Vietnamese settings. Although overall trends are similar, performance on Vietnamese prompts shows less stability across steps and models, particularly at Step~3. In several cases, English prompts yield more stable and slightly higher F1 scores across reasoning stages, while Vietnamese performance fluctuates more. This suggests that hallucination detection is more sensitive to Vietnamese prompt formulation, especially for fine-grained attribute descriptions.

\subsubsection{Vulnerability to Plausible Rationales} \label{sub:Case_Study}

\paragraph{\hyperref[rq:rationale]{RQ2}.}
When a textual rationale is introduced, hallucination detection is no longer driven purely by visual evidence; instead, the additional language steers the consistency judgment, weakening image grounding across models (Table~\ref{tab:add_expl_no_expl_cols}). This effect is dominated by a sharp degradation in recall, indicating that models increasingly accept incorrect statements as image-consistent when a rationale is present.

Clinician-written rationales are generally more harmful than LLM-generated ones. Their clinically standard, fluent, and report-like style makes them more persuasive to the model’s language prior, leading to missed hallucinations.
This is evident for Gemini~2.5~Pro and Gemini~2.5~Flash, where F1 decreases from 0.67/0.69 to 0.57/0.57 under clinician rationales, or an F1 drop from 0.60 to 0.53 for MedM-VL 3D. 
In contrast, LLM-generated rationales do not induce a consistent degradation. Gemini 2.5 Pro and Gemini 2.5 Flash maintain high recall even under LLM rationales (0.96 and 0.98), achieving F1 scores up to 0.69. This robustness likely stems from LLM rationales being less tightly constrained by real clinical practice, often appearing generic or insufficiently persuasive to override visual evidence.
Med-Flamingo is an exception, showing strong sensitivity even to LLM rationales, however the model still suffered a major decrease in hallucination detection capabilities. 
Overall, these results reveal a persuasion-over-evidence failure mode, where clinically plausible narratives override visual verification, exposing a weakness in visual grounding.

\subsubsection{Effects of Prior and External Knowledge}
\paragraph{\hyperref[rq:knowledge]{RQ3}.}
We also observe that knowledge injection for both types of knowledge only yields marginal F1 changes for most models in most steps, and does not help to alleviate hallucination meaningfully (Table~\ref{tab:internal_vs_external}). Step 2 generally yields small F1 gains or remains unchanged, indicating that contextual constraints align well with lesion presence verification, though occasional degradations (e.g., LLaVA-Med) suggest limited robustness. In contrast, Step 3 shows little improvement and frequent drops, highlighting a persistent bottleneck where lightweight prompts fail to support fine-grained attribute grounding and models revert to textual priors. Step 4 typically improves or stays flat, but gains are largely driven by a precision–recall trade-off, with higher sensitivity accompanied by increased false positives. Overall, intermediate knowledge primarily redistributes errors rather than achieving consistent hallucination reduction, benefiting lesion identification more than feature characterization.

\begin{table}[h]
\centering
\tiny
\setlength{\tabcolsep}{1.6pt}
\renewcommand{\arraystretch}{1.1}

\begin{adjustbox}{width=0.9\columnwidth}
\begin{tabular}{l| cc| ccc |ccc| ccc}
\toprule
\textbf{Model} &
\multicolumn{2}{c|}{\textbf{Knowledge}} &
\multicolumn{3}{c|}{\textbf{Step 2}} &
\multicolumn{3}{c|}{\textbf{Step 3}} &
\multicolumn{3}{c}{\textbf{Step 4}} \\
\midrule
& \textbf{Prior} & \textbf{Ext} &
\textbf{P} & \textbf{R} & \textbf{F1} &
\textbf{P} & \textbf{R} & \textbf{F1} &
\textbf{P} & \textbf{R} & \textbf{F1} \\
\midrule
\multirow{4}{*}{\makecell[c]{Qwen3\\VL-4B}}
&  & 
& 0.58 & 0.74 & 0.65
& 0.56 & 0.70 & 0.62
& 0.62 & 0.65 & 0.63 \\
& \cmark & 
& 0.53 & \textbf{0.94} & 0.68
& 0.58 & 0.72 & 0.64
& \textbf{0.63} & 0.83 & \textbf{0.72} \\
&  & \cmark
& \textbf{0.60} & 0.88 & \textbf{0.71}
& 0.54 & 0.78 & 0.64
& 0.55 & 0.50 & 0.52 \\
& \cmark & \cmark
& 0.59 & 0.84 & 0.69
& \textbf{0.59} & 0.78 & \textbf{0.67}
& 0.55 & \textbf{0.88} & 0.67 \\
\midrule
\multirow{4}{*}{\makecell[c]{Med\\Flamingo}}
&  &
& 0.63 & 0.69 & 0.66
& \textbf{0.51} & 0.82 & 0.63
& 0.53 & 0.67 & 0.59 \\
& \cmark &
& 0.55 & 0.94 & 0.70
& 0.50 & \textbf{0.88} & \textbf{0.64}
& 0.54 & \textbf{0.79} & \textbf{0.64} \\
&  & \cmark
& 0.55 & \textbf{0.96} & \textbf{0.70}
& 0.47 & 0.80 & 0.59
& 0.52 & 0.77 & 0.62 \\
& \cmark & \cmark
& \textbf{0.65} & 0.74 & 0.69
& 0.47 & 0.82 & 0.59
& \textbf{0.55} & 0.71 & 0.62 \\
\midrule
\multirow{4}{*}{\makecell[c]{Gemini\\ 2.5\\Flash}}
&  &
& 0.65 & 0.82 & \textbf{0.72}
& 0.56 & \textbf{0.84} & \textbf{0.67}
& \textbf{0.64} & 0.58 & 0.61 \\
& \cmark &
& 0.52 & 0.65 & 0.58
& 0.50 & 0.58 & 0.54
& 0.62 & 0.63 & 0.63 \\
&  & \cmark
& 0.63 & \textbf{0.85} & \textbf{0.72}
& 0.57 & 0.82 & \textbf{0.67}
& 0.60 & 0.72 & \textbf{0.65} \\
& \cmark & \cmark
& \textbf{0.66} & 0.80 & \textbf{0.72}
& \textbf{0.58} & 0.78 & \textbf{0.67}
& 0.57 & \textbf{0.77} & \textbf{0.65} \\

\bottomrule
\end{tabular}
\end{adjustbox}

\caption{Different knowledge input across steps. Prior/Ext indicate whether prior knowledge or external knowledge is provided (both checked means +Both; none checked means no knowledge).}
\label{tab:internal_vs_external}
\vspace{-10pt}
\end{table}

\begin{table}[h]
\centering
\large
\setlength{\tabcolsep}{5.2pt}
\renewcommand{\arraystretch}{1.2}

\begin{adjustbox}{width=\columnwidth}
\begin{tabular}{l !{\vrule width 0.8pt} c !{\vrule width 0.8pt}
cc !{\vrule width 0.8pt}
cc !{\vrule width 0.8pt}
cc !{\vrule width 0.8pt}
cc}
\toprule
\multirow{2}{*}{\textbf{Model}} & \multirow{2}{*}{\textbf{Input}} &
\multicolumn{2}{c!{\vrule width 0.8pt}}{\textbf{Step 1}} &
\multicolumn{2}{c!{\vrule width 0.8pt}}{\textbf{Step 2}} &
\multicolumn{2}{c!{\vrule width 0.8pt}}{\textbf{Step 3}} &
\multicolumn{2}{c}{\textbf{Step 4}} \\
\cmidrule(lr){3-4}\cmidrule(lr){5-6}\cmidrule(lr){7-8}\cmidrule(lr){9-10}
& & \textbf{VI} & \textbf{EN} & \textbf{VI} & \textbf{EN} & \textbf{VI} & \textbf{EN} & \textbf{VI} & \textbf{EN} \\
\midrule

\multirow{2}{*}{\makecell[l]{Med\\Flamingo}}
& PET + CT
& -- & 0.64 & -- & 0.64 & -- & \textbf{0.66} & -- & \textbf{0.66} \\
& PET/CT
& -- & \textbf{0.77}\textcolor[HTML]{008000}{$\uparrow$} 
& -- & \textbf{0.66}\textcolor[HTML]{008000}{$\uparrow$} 
& -- & 0.63\textcolor[HTML]{FF0000}{$\downarrow$} 
& -- & 0.59\textcolor[HTML]{FF0000}{$\downarrow$} \\
\midrule

\multirow{2}{*}{\makecell[l]{Gemini 2.5\\Flash}}
& PET + CT
& 0.79 & 0.81 & 0.68 & \textbf{0.72} & 0.62 & 0.62 & 0.51 & 0.52 \\
& PET/CT
& \textbf{0.98}\textcolor[HTML]{008000}{$\uparrow$} 
& \textbf{0.98}\textcolor[HTML]{008000}{$\uparrow$} 
& \textbf{0.72}\textcolor[HTML]{008000}{$\uparrow$} 
& 0.71\textcolor[HTML]{FF0000}{$\downarrow$} 
& \textbf{0.67}\textcolor[HTML]{008000}{$\uparrow$} 
& \textbf{0.67}\textcolor[HTML]{008000}{$\uparrow$} 
& \textbf{0.61}\textcolor[HTML]{008000}{$\uparrow$} 
& \textbf{0.62}\textcolor[HTML]{008000}{$\uparrow$} \\
\midrule

\multirow{2}{*}{\makecell[l]{Qwen3\\VL-4B}}
& PET + CT
& 0.60 & 0.79 & 0.31 & 0.34 & 0.29 & 0.20 & \textbf{0.67} & \textbf{0.64} \\
& PET/CT
& \textbf{0.92}\textcolor[HTML]{008000}{$\uparrow$} 
& \textbf{0.89}\textcolor[HTML]{008000}{$\uparrow$} 
& \textbf{0.68}\textcolor[HTML]{008000}{$\uparrow$} 
& \textbf{0.65}\textcolor[HTML]{008000}{$\uparrow$} 
& \textbf{0.66}\textcolor[HTML]{008000}{$\uparrow$} 
& \textbf{0.62}\textcolor[HTML]{008000}{$\uparrow$} 
& 0.63\textcolor[HTML]{FF0000}{$\downarrow$} 
& 0.62\textcolor[HTML]{FF0000}{$\downarrow$} \\
\bottomrule
\end{tabular}
\end{adjustbox}
\caption{Comparison of F1-score between PET+CT (separate) and fused PET/CT across steps in both English and Vietnamese.}
\label{tab:pet_ct_vs_petct}
\vspace{-10pt}
\end{table}

\subsubsection{Clinically Aligned PET/CT Fusion}
\paragraph{\hyperref[rq:fusion]{RQ4.}}

As shown in Table~\ref{tab:pet_ct_vs_petct}, clinically fused PET/CT inputs consistently outperform simple PET+CT concatenation across models. For Gemini~2.5~Flash, PET/CT fusion yields a large gain at Step~1 (VI), improving F1 from 0.79 to 0.98, and also improves fine-grained reasoning at Step~3 (EN) from 0.62 to 0.67. Similarly, Qwen3-VL-4B benefits substantially from PET/CT fusion, with Step~3 (EN) F1 increasing from 0.20 to 0.63, more than a threefold improvement. These results indicate that clinically grounded PET/CT fusion significantly enhances cross-modal alignment and fine-grained reasoning compared to naïve modality concatenation.


\vspace{-5pt}
\section{Conclusion}

This paper introduced \textit{Med-StepBench}, a large-scale step-wise hallucination detection benchmark for 2D and 3D oncological PET/CT, comprising over 1 million image–statement pairs across four clinically grounded diagnostic stages. Med-StepBench enabled detailed examination of whether model predictions were supported by correct anatomical localization, lesion identification, feature characterization, and diagnostic reasoning. Through extensive experiments on general-purpose and medical vision language models, we showed that hallucinations emerged and propagated differently across reasoning stages, and that current models were particularly susceptible to clinically plausible but incorrect intermediate explanations. Our results highlighted critical limitations in visual grounding and multi-step clinical reasoning, positioning \textit{Med-StepBench} as a practical benchmark for advancing safer and more reliable medical vision language models.

\appendix

\section*{Ethical Statement}

This study was conducted in accordance with established medical research ethics standards and institutional regulations. All imaging data were retrospectively collected and fully de-identified under approval of the Institutional Review Board (IRB), with a formally granted waiver of informed consent due to minimal risk. Personal protected information (PPI) was removed following international privacy standards (e.g., HIPAA Safe Harbor). 
The dataset will be publicly released for non-commercial research purposes under specified usage terms upon paper acceptance. 

\section*{Acknowledgements}
This research is partly supported by Toray Industries (H.K.) Vietnam Co., Ltd.
This research is funded by Hanoi University of Science and Technology (HUST) under grant number  T2024-TĐ-002.


\begin{thebibliography}{}

\bibitem[\protect\citeauthoryear{Asgari \bgroup \em et al.\egroup }{2025}]{asgari2025framework}
Elham Asgari, Nina Monta{\~n}a-Brown, Magda Dubois, Saleh Khalil, Jasmine Balloch, Joshua~Au Yeung, and Dominic Pimenta.
\newblock A framework to assess clinical safety and hallucination rates of llms for medical text summarisation.
\newblock {\em npj Digital Medicine}, 8(1):274, 2025.

\bibitem[\protect\citeauthoryear{Bae \bgroup \em et al.\egroup }{2023}]{mimic-cxr-vqa}
Seongsu Bae, Daeun Kyung, Jaehee Ryu, Eunbyeol Cho, Gyubok Lee, Sunjun Kweon, Jungwoo Oh, Lei Ji, Eric Chang, Tackeun Kim, et~al.
\newblock Ehrxqa: A multi-modal question answering dataset for electronic health records with chest x-ray images.
\newblock {\em Advances in Neural Information Processing Systems}, 36:3867--3880, 2023.

\bibitem[\protect\citeauthoryear{Bai \bgroup \em et al.\egroup }{2024}]{M3D2024}
Fan Bai, Yuxin Du, Tiejun Huang, Max~Qinghu Meng, and Bo~Zhao.
\newblock M3d: Advancing 3d medical image analysis with multi-modal large language models, 2024.

\bibitem[\protect\citeauthoryear{Bai \bgroup \em et al.\egroup }{2025}]{Qwen3VLTechReport2025}
Shuai Bai, Yuxuan Cai, Ruizhe Chen, et~al.
\newblock Qwen3-vl technical report, 2025.

\bibitem[\protect\citeauthoryear{Ben~Abacha \bgroup \em et al.\egroup }{2021}]{vqamed2021}
Asma Ben~Abacha, Mourad Sarrouti, Dina Demner-Fushman, Sadid~A. Hasan, and Henning M{\"u}ller.
\newblock Overview of the vqa-med task at imageclef 2021: Visual question answering and generation in the medical domain.
\newblock In {\em CLEF 2021 Conference and Labs of the Evaluation Forum - Working Notes}, 2021.

\bibitem[\protect\citeauthoryear{Chang \bgroup \em et al.\egroup }{2025}]{MedHEval2025}
Aofei Chang, Le~Huang, Parminder Bhatia, Taha Kass-Hout, Fenglong Ma, and Cao Xiao.
\newblock Medheval: Benchmarking hallucinations and mitigation strategies in medical large vision--language models, 2025.

\bibitem[\protect\citeauthoryear{Chen \bgroup \em et al.\egroup }{2024}]{chen2025detecting}
Jiawei Chen, Dingkang Yang, Tong Wu, Yue Jiang, Xiaolu Hou, Mingcheng Li, Shunli Wang, Dongling Xiao, Ke~Li, and Lihua Zhang.
\newblock Detecting and evaluating medical hallucinations in large vision language models, 2024.
\newblock Med-HallMark benchmark.

\bibitem[\protect\citeauthoryear{Comanici \bgroup \em et al.\egroup }{2025}]{Gemini25Report2025}
Gheorghe Comanici, Eric Bieber, Mike Schaekermann, et~al.
\newblock Gemini 2.5: Pushing the frontier with advanced reasoning, multimodality, long context, and next generation agentic capabilities, 2025.

\bibitem[\protect\citeauthoryear{Gu \bgroup \em et al.\egroup }{2026}]{medvh}
Zishan Gu, Jiayuan Chen, Fenglin Liu, Changchang Yin, and Ping Zhang.
\newblock Medvh: Toward systematic evaluation of hallucination for large vision language models in the medical context.
\newblock {\em Advanced Intelligent Systems}, 8(1):2500255, 2026.

\bibitem[\protect\citeauthoryear{He \bgroup \em et al.\egroup }{2020}]{He2020PathVQA3Q}
Xuehai He, Yichen Zhang, Luntian Mou, Eric~P. Xing, and Pengtao Xie.
\newblock Pathvqa: 30000+ questions for medical visual question answering.
\newblock {\em arXiv preprint}, 2020.

\bibitem[\protect\citeauthoryear{Hofman and Hicks}{2016}]{hofman2016how}
Michael~S. Hofman and Rodney~J. Hicks.
\newblock How we read oncologic {FDG} {PET}/{CT}.
\newblock {\em Cancer Imaging}, 16(1):35, 2016.

\bibitem[\protect\citeauthoryear{Hou \bgroup \em et al.\egroup }{2025}]{ijcai2025p824}
Xiaodi Hou, Xiaobo Li, Mingyu Lu, Simiao Wang, and Yijia Zhang.
\newblock Rrg-mamba: Efficient radiology report generation with state space model.
\newblock In James Kwok, editor, {\em Proceedings of the Thirty-Fourth International Joint Conference on Artificial Intelligence, {IJCAI-25}}, pages 7410--7418. International Joint Conferences on Artificial Intelligence Organization, 8 2025.
\newblock Main Track.

\bibitem[\protect\citeauthoryear{Jiang \bgroup \em et al.\egroup }{2025}]{ijcai2025p46}
Songtao Jiang, Yan Zhang, Ruizhe Chen, Tianxiang Hu, Yeying Jin, Qinglin He, Yang Feng, Jian Wu, and Zuozhu Liu.
\newblock Modality-fair preference optimization for trustworthy mllm alignment.
\newblock In James Kwok, editor, {\em Proceedings of the Thirty-Fourth International Joint Conference on Artificial Intelligence, {IJCAI-25}}, pages 403--411. International Joint Conferences on Artificial Intelligence Organization, 8 2025.
\newblock Main Track.

\bibitem[\protect\citeauthoryear{Kim \bgroup \em et al.\egroup }{2025}]{kim2025medical}
Yubin Kim, Hyewon Jeong, Shan Chen, Shuyue~Stella Li, Chanwoo Park, Mingyu Lu, Kumail Alhamoud, Jimin Mun, Cristina Grau, Minseok Jung, et~al.
\newblock Medical hallucinations in foundation models and their impact on healthcare.
\newblock {\em arXiv preprint arXiv:2503.05777}, 2025.

\bibitem[\protect\citeauthoryear{Lau \bgroup \em et al.\egroup }{2018}]{vqa-rad}
Jason~J Lau, Soumya Gayen, Asma Ben~Abacha, and Dina Demner-Fushman.
\newblock A dataset of clinically generated visual questions and answers about radiology images.
\newblock {\em Scientific data}, 5(1):1--10, 2018.

\bibitem[\protect\citeauthoryear{Li \bgroup \em et al.\egroup }{2023}]{LLaVAMed2023}
Chunyuan Li, Cliff Wong, Sheng Zhang, Naoto Usuyama, Haotian Liu, Jianwei Yang, Tristan Naumann, Hoifung Poon, and Jianfeng Gao.
\newblock Llava-med: Training a large language-and-vision assistant for biomedicine in one day, 2023.

\bibitem[\protect\citeauthoryear{Liu \bgroup \em et al.\egroup }{2021}]{Liu2021SlakeAS}
Bo~Liu, Li-Ming Zhan, Li~Xu, Lin Ma, Yan Yang, and Xiao-Ming Wu.
\newblock Slake: A semantically-labeled knowledge-enhanced dataset for medical visual question answering.
\newblock In {\em 2021 IEEE 18th international symposium on biomedical imaging (ISBI)}, pages 1650--1654. IEEE, 2021.

\bibitem[\protect\citeauthoryear{Liu \bgroup \em et al.\egroup }{2023}]{LLaVA2023}
Haotian Liu, Chunyuan Li, Qingyang Wu, and Yong~Jae Lee.
\newblock Visual instruction tuning, 2023.

\bibitem[\protect\citeauthoryear{Liu \bgroup \em et al.\egroup }{2025}]{liu2025argus}
Che Liu, Zhongwei Wan, Yuqi Wang, Hui Shen, Haozhe Wang, Kangyu Zheng, Mi~Zhang, and Rossella Arcucci.
\newblock Argus: benchmarking and enhancing vision-language models for 3d radiology report generation.
\newblock In {\em Findings of the Association for Computational Linguistics: ACL 2025}, pages 16448--16460, 2025.

\bibitem[\protect\citeauthoryear{Moor \bgroup \em et al.\egroup }{2023}]{MedFlamingoModel}
Michael Moor, Qian Huang, Shirley Wu, Michihiro Yasunaga, Yash Dalmia, Jure Leskovec, Cyril Zakka, Eduardo~Pontes Reis, and Pranav Rajpurkar.
\newblock Med-flamingo: a multimodal medical few-shot learner.
\newblock In {\em Machine Learning for Health (ML4H)}, pages 353--367. PMLR, 2023.

\bibitem[\protect\citeauthoryear{Nguyen \bgroup \em et al.\egroup }{2025a}]{HEALMedVQA2025}
Dung Nguyen, Minh~Khoi Ho, Huy Ta, Thanh~Tam Nguyen, Qi~Chen, Kumar Rav, Quy~Duong Dang, Satwik Ramchandre, Son~Lam Phung, Zhibin Liao, Minh-Son To, Johan Verjans, Phi~Le Nguyen, and Vu~Minh~Hieu Phan.
\newblock Localizing before answering: A benchmark for grounded medical visual question answering.
\newblock In James Kwok, editor, {\em Proceedings of the Thirty-Fourth International Joint Conference on Artificial Intelligence, {IJCAI-25}}, pages 7670--7678. International Joint Conferences on Artificial Intelligence Organization, 8 2025.
\newblock Main Track.

\bibitem[\protect\citeauthoryear{Nguyen \bgroup \em et al.\egroup }{2025b}]{ViMedPET2025}
Huu~Tien Nguyen, Dac~Thai Nguyen, Duc Nguyen~The Minh, Trung~Thanh Nguyen, Thao~Nguyen Truong, Hieu Pham, Johan Barthelemy, Tran~Minh Quan, Quoc Viet~Hung Nguyen, Thanh~Tam Nguyen, Mai~Hong Son, Chau~Quynh Anh, Thanh~Trung Nguyen, and Phi~Le Nguyen.
\newblock Toward a vision-language foundation model for medical data: Multimodal dataset and benchmarks for vietnamese {PET}/{CT} report generation.
\newblock In {\em The Thirty-ninth Annual Conference on Neural Information Processing Systems Datasets and Benchmarks Track}, 2025.

\bibitem[\protect\citeauthoryear{OpenAI}{2024}]{GPT4oTechReport2025}
OpenAI.
\newblock Gpt-4o system card, 2024.

\bibitem[\protect\citeauthoryear{Shi \bgroup \em et al.\egroup }{2025}]{MedMVL2025}
Yiming Shi, Shaoshuai Yang, Xun Zhu, Haoyu Wang, Xiangling Fu, Miao Li, and Ji~Wu.
\newblock Medm-vl: What makes a good medical lvlm?, 2025.

\bibitem[\protect\citeauthoryear{Sun \bgroup \em et al.\egroup }{2025}]{sun2025understanding}
Xiaoxi Sun, Jianxin Liang, Yueqian Wang, Huishuai Zhang, and Dongyan Zhao.
\newblock Understanding visual detail hallucinations of large vision-language models.
\newblock In {\em Proceedings of the Thirty-Fourth International Joint Conference on Artificial Intelligence}, pages 1900--1908, 2025.

\bibitem[\protect\citeauthoryear{Wu \bgroup \em et al.\egroup }{2023}]{wu2023medklip}
Chaoyi Wu, Xiaoman Zhang, Ya~Zhang, Yanfeng Wang, and Weidi Xie.
\newblock Medklip: Medical knowledge enhanced language-image pre-training for x-ray diagnosis.
\newblock In {\em Proceedings of the IEEE/CVF international conference on computer vision}, pages 21372--21383, 2023.

\bibitem[\protect\citeauthoryear{Wu \bgroup \em et al.\egroup }{2024}]{wu2024hallucination}
Jinge Wu, Yunsoo Kim, and Honghan Wu.
\newblock Hallucination benchmark in medical visual question answering.
\newblock In {\em The Second Tiny Papers Track at ICLR 2024}, 2024.

\bibitem[\protect\citeauthoryear{Xia \bgroup \em et al.\egroup }{2024}]{CARES2024}
Peng Xia, Ze~Chen, Juanxi Tian, Yangrui Gong, Ruibo Hou, Yue Xu, Zhenbang Wu, Zhiyuan Fan, Yiyang Zhou, Kangyu Zhu, et~al.
\newblock Cares: A comprehensive benchmark of trustworthiness in medical vision language models.
\newblock {\em Advances in Neural Information Processing Systems}, 37:140334--140365, 2024.

\bibitem[\protect\citeauthoryear{Yan \bgroup \em et al.\egroup }{2025}]{yan2025worse}
Qianqi Yan, Xuehai He, Xiang Yue, and Xin~Eric Wang.
\newblock Worse than random? an embarrassingly simple probing evaluation of large multimodal models in medical vqa.
\newblock In {\em Findings of the Association for Computational Linguistics: ACL 2025}, pages 19188--19205, 2025.

\bibitem[\protect\citeauthoryear{Zhang \bgroup \em et al.\egroup }{2023}]{pmc-vqa}
Xiaoman Zhang, Chaoyi Wu, Ziheng Zhao, Weixiong Lin, Ya~Zhang, Yanfeng Wang, and Weidi Xie.
\newblock Pmc-vqa: Visual instruction tuning for medical visual question answering.
\newblock {\em arXiv preprint}, 2023.

\bibitem[\protect\citeauthoryear{Zhou \bgroup \em et al.\egroup }{2025}]{drvd-bench}
Tianhong Zhou, Yin Xu, Yingtao Zhu, Chuxi Xiao, Haiyang Bian, Lei Wei, and Xuegong Zhang.
\newblock Dr{VD}-bench: Do vision-language models reason like human doctors in medical image diagnosis?
\newblock In {\em The Thirty-ninth Annual Conference on Neural Information Processing Systems Datasets and Benchmarks Track}, 2025.

\bibitem[\protect\citeauthoryear{Zhu \bgroup \em et al.\egroup }{2025}]{zhu2025can}
Zhihong Zhu, Yunyan Zhang, Xianwei Zhuang, Fan Zhang, Zhongwei Wan, Yuyan Chen, Qingqing Long, Yefeng Zheng, and Xian Wu.
\newblock Can we trust {AI} doctors? a survey of medical hallucination in large language and large vision-language models.
\newblock In Wanxiang Che, Joyce Nabende, Ekaterina Shutova, and Mohammad~Taher Pilehvar, editors, {\em Findings of the Association for Computational Linguistics: ACL 2025}, pages 6748--6769, Vienna, Austria, July 2025. Association for Computational Linguistics.

\bibitem[\protect\citeauthoryear{Zuo and Jiang}{2025}]{zuo2025medhallbench}
Kaiwen Zuo and Yirui Jiang.
\newblock Medhallbench: A new benchmark for assessing hallucination in medical large language models.
\newblock In {\em AAAI Bridge Program on AI for Medicine and Healthcare}, pages 205--213. PMLR, 2025.

\end{thebibliography}
\end{document}